%% file: aaai24.tex
\documentclass[letterpaper]{article} 
\usepackage{aaai24}  
\usepackage{times}  
\usepackage{helvet}  
\usepackage{courier}  
\usepackage[hyphens]{url}  
\usepackage{graphicx} 
\usepackage{natbib}  
\usepackage{caption} 
\usepackage{algorithm}
\usepackage{algorithmic}

\usepackage{newfloat}
\usepackage{listings}

\usepackage{subcaption} 
\usepackage{multirow}
\usepackage{booktabs}

\DeclareCaptionStyle{ruled}{labelfont=normalfont,labelsep=colon,strut=off} 
\floatstyle{ruled}
\newfloat{listing}{tb}{lst}{}
\floatname{listing}{Listing}
\title{SkyScript: A Large and Semantically Diverse Vision-Language Dataset for Remote Sensing}
\author{
    Zhecheng Wang\textsuperscript{\rm 1},
    Rajanie Prabha\textsuperscript{\rm 1}\equalcontrib,
    Tianyuan Huang\textsuperscript{\rm 1}\equalcontrib,
    Jiajun Wu\textsuperscript{\rm 1},
    Ram Rajagopal\textsuperscript{\rm 1}
}
\affiliations {
    \textsuperscript{\rm 1}Stanford University\\
    \{zhecheng, rajanie, tianyuah, ramr\}@stanford.edu, jiajunwu@cs.stanford.edu
}

\usepackage{bibentry}

\nocopyright

\begin{document}

\maketitle

\begin{abstract}
Remote sensing imagery, despite its broad applications in helping achieve Sustainable Development Goals and tackle climate change, has not yet benefited from the recent advancements of versatile, task-agnostic vision language models (VLMs).
A key reason is that the large-scale, semantically diverse image-text dataset required for developing VLMs is still absent for remote sensing images.
Unlike natural images, remote sensing images and their associated text descriptions cannot be efficiently collected from the public Internet at scale. 
In this work, we bridge this gap by using geo-coordinates to automatically connect open, unlabeled remote sensing images with rich semantics covered in OpenStreetMap, and thus construct SkyScript, a comprehensive vision-language dataset for remote sensing images, comprising 2.6 million image-text pairs covering 29K distinct semantic tags. 
With continual pre-training on this dataset, we obtain a VLM that surpasses baseline models with a 6.2\% average accuracy gain in zero-shot scene classification across seven benchmark datasets. 
It also demonstrates the ability of zero-shot transfer for fine-grained object attribute classification and cross-modal retrieval.
We hope this dataset can support the advancement of VLMs for various multi-modal tasks in remote sensing, such as open-vocabulary classification, retrieval, captioning, and text-to-image synthesis.

\end{abstract}

\input{introduction}
\input{related_work}

\input{dataset}

\input{experiments}

\input{conclusion}
\input{ackowledgements}

\appendix

\bibliography{aaai24}

\clearpage

\appendix

\renewcommand\thefigure{A\arabic{figure}}
\renewcommand\thetable{A\arabic{table}}

\setcounter{figure}{0}
\setcounter{table}{0}

\section{Appendix}

\input{appendix}

\end{document}

%% file: introduction.tex
\section{Introduction}

Remote sensing imagery plays an important role for achieving Sustainable Development Goals (SDGs). 
Applying computer vision on remote sensing images can automate a broad range of applications, such as poverty estimation \cite{Jean2016CombiningSI}, crop yield prediction \cite{You2017DeepGP}, deforestation detection \cite{Torres2021DeforestationDW}, and renewable energy mapping \cite{Yu2018DeepSolarAM,Kruitwagen2021AGI}.
Facing the increasing risk of climate change, remote sensing imagery and the vision models built for it can further contribute to both mitigation and adaptation by enabling the observation of the Earth surface \cite{helber2019eurosat}, detection of high-pollution industry \cite{Lee2021ScalableDL}, evaluation of carbon stock \cite{Reiersen2022ReforesTreeAD}, and identification of vulnerable infrastructure and populations 
\cite{Huang2021GridTracerAM}.

\begin{figure}[t]
\centering
\includegraphics[width=\columnwidth]{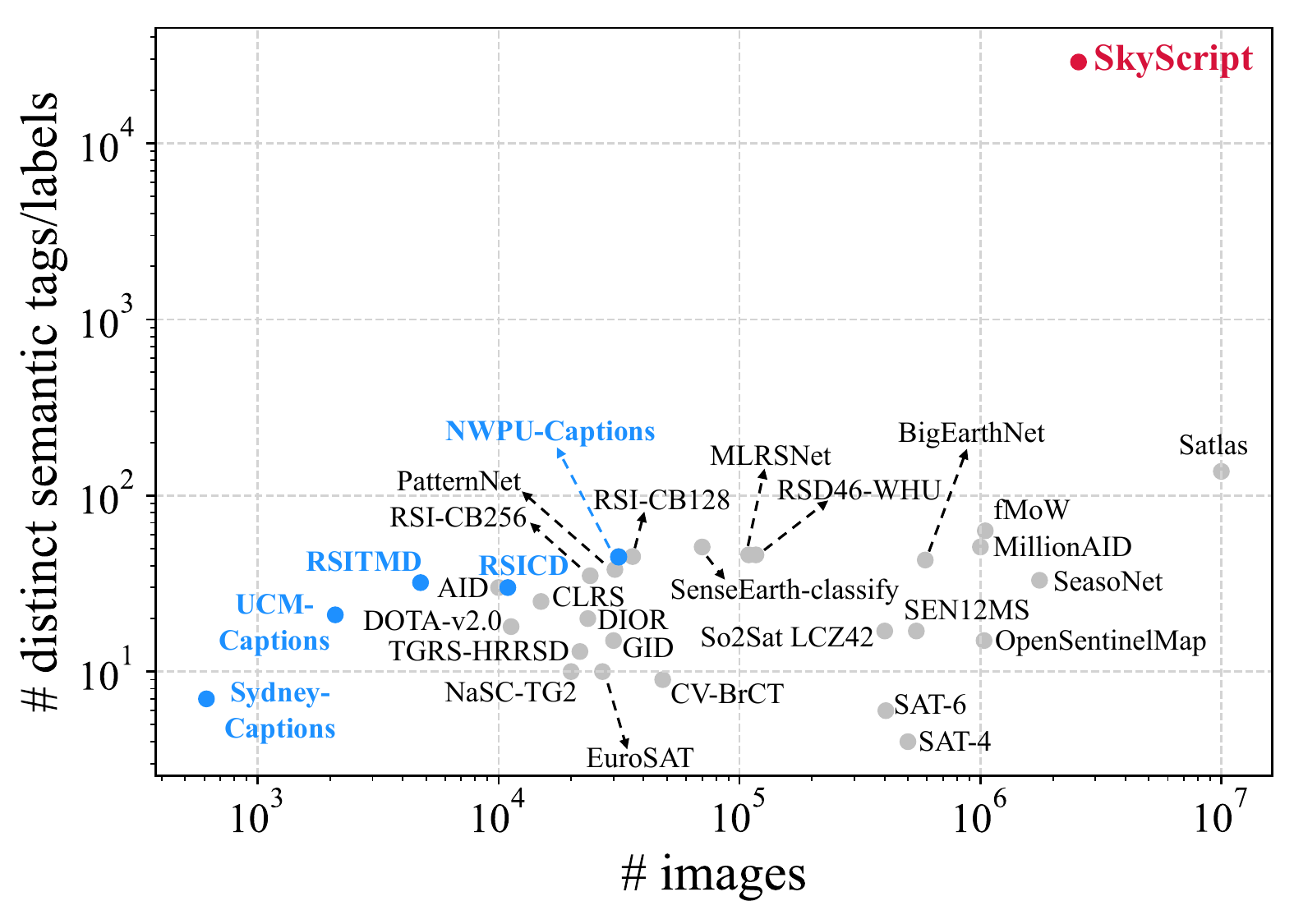} 
\caption{Comparison of general-purpose remote sensing datasets. Grey circles denote datasets for classification, object detection, or semantic segmentation (only datasets with $\geq$10,000 images are shown). 
Blue circles denote image-text datasets. 
Our dataset, SkyScript, is over two orders of magnitude more semantically diverse than existing remote sensing image-text datasets.}
\label{fig:comparison}
\end{figure}

Despite numerous task-specific supervised learning models developed for remote sensing images, this domain has not yet fully benefited from the recent advancements of task-agnostic, versatile vision language models (VLMs) such as CLIP \cite{radford2021learning,jia2021scaling,alayrac2022flamingo}. 
This is because a key ingredient in the VLM's recipe for gaining versatility and generalizability is the large and semantically diverse collection of image-text pairs, which is still not readily available for remote sensing images. 
In the recent development of VLMs, such image-text pairs, ranging from million to billion scale, are usually collected from the public Internet through web crawling \cite{schuhmann2022laion}. 
By contrast, remote sensing images are collected and owned exclusively by Earth observation companies, government agencies, or intergovernmental organizations (e.g., European Space Agency). 
Although these imagery data may be accessed through specialized, often non-free data pipelines (e.g., paid API service), they cannot be crawled from the web at scale.
Even when these images can be obtained, they are usually standalone—not like web images which are often surrounded by semantically relevant text.

Due to the domain familiarity required for annotating remote sensing images, obtaining large and semantically diverse data through human annotations is also challenging. 
As Figure \ref{fig:comparison} shows, the sizes of existing remote sensing datasets can rarely reach the million-level, and the semantic classes are no more than 150. This significantly constrains the development of VLMs in the remote sensing domain. 

In this work, we bridge this gap by constructing SkyScript\footnote{The dataset and associated models are publicly available at \url{https://github.com/wangzhecheng/SkyScript}}, a large and semantically diverse image-text dataset for remote sensing.
We achieve this by using geo-coordinates to connect open, unlabeled remote sensing images on Google Earth Engine (GEE) with rich semantic information covered in the OpenStreetMap (OSM) database. 
This yields a dataset with global coverage containing 2.6 million image-text pairs and covering 29K distinct semantic tags—two orders of magnitude richer than the existing remote sensing image-text datasets (Figure \ref{fig:comparison}). 
The semantic information spans across object categories, subcategories, and detailed attributes (e.g., road surface materials). 
We demonstrate the value of this dataset by using continual pre-training to obtain a CLIP model that substantially outperforms the original CLIP and other baselines on three downstream remote sensing tasks: zero-shot scene classification, fined-grained attribute classification, and cross-modal retrieval. 
Our major contributions are summarized as follows:

\begin{itemize}
    \item We create SkyScript, a large-scale, semantically diverse vision-language dataset for remote sensing images.
    
    \item SkyScript enables the development of a CLIP model for remote sensing that outperforms the original CLIP and other baselines in zero-shot scene classification.
    
    \item We further demonstrate the capabilities of SkyScript and its derived models on remote sensing cross-modal retrieval and zero-shot fine-grained attribute classification.
\end{itemize}

\begin{figure*}[t]
  \centering
  \begin{subfigure}{0.34\textwidth}
    \centering
   \includegraphics[width=\textwidth]{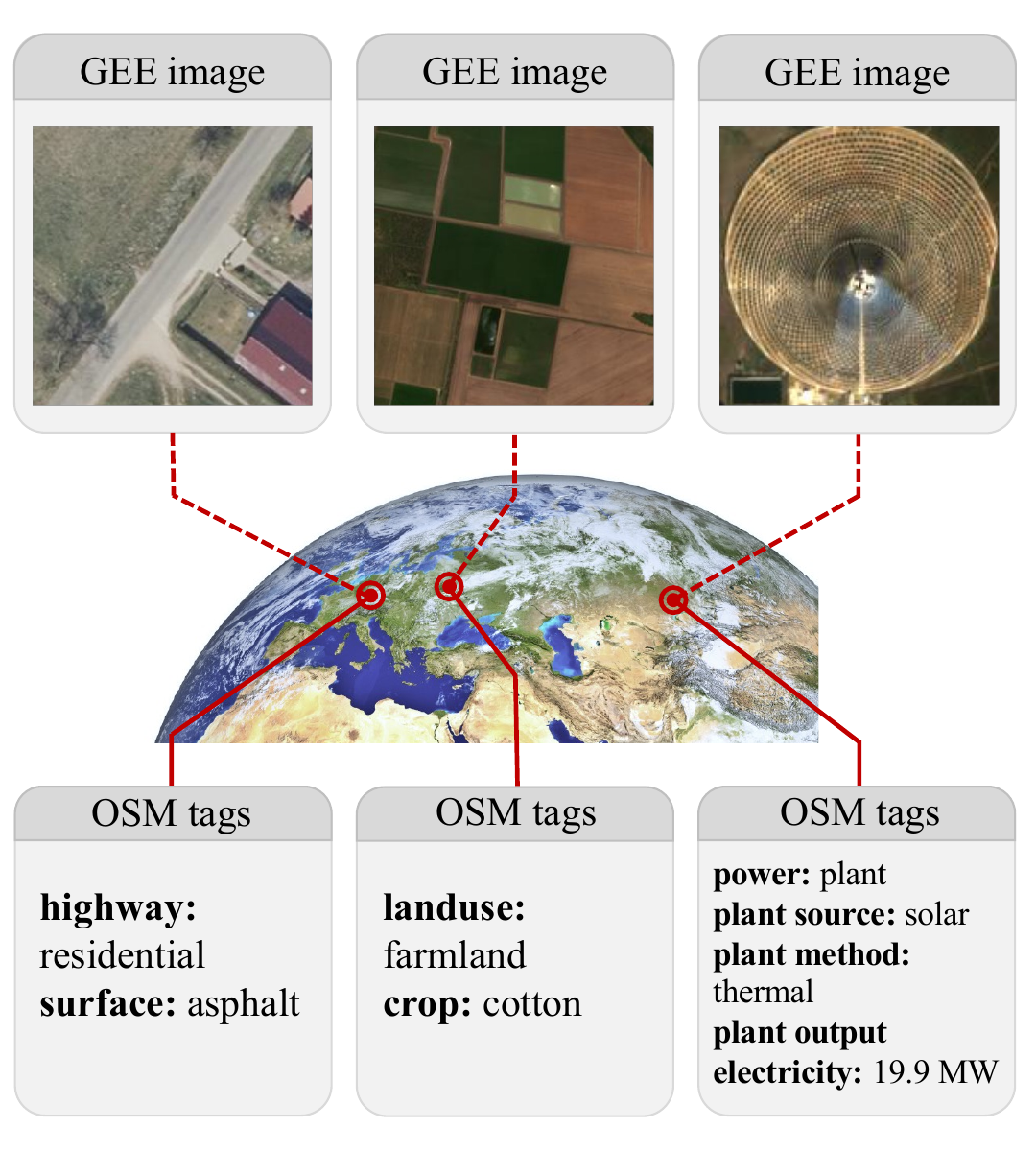}
    \caption{}
    \label{fig:image_text_register}
  \end{subfigure}%
  \hfill
  \begin{subfigure}{0.655\textwidth}
    \centering
   \includegraphics[width=\textwidth]{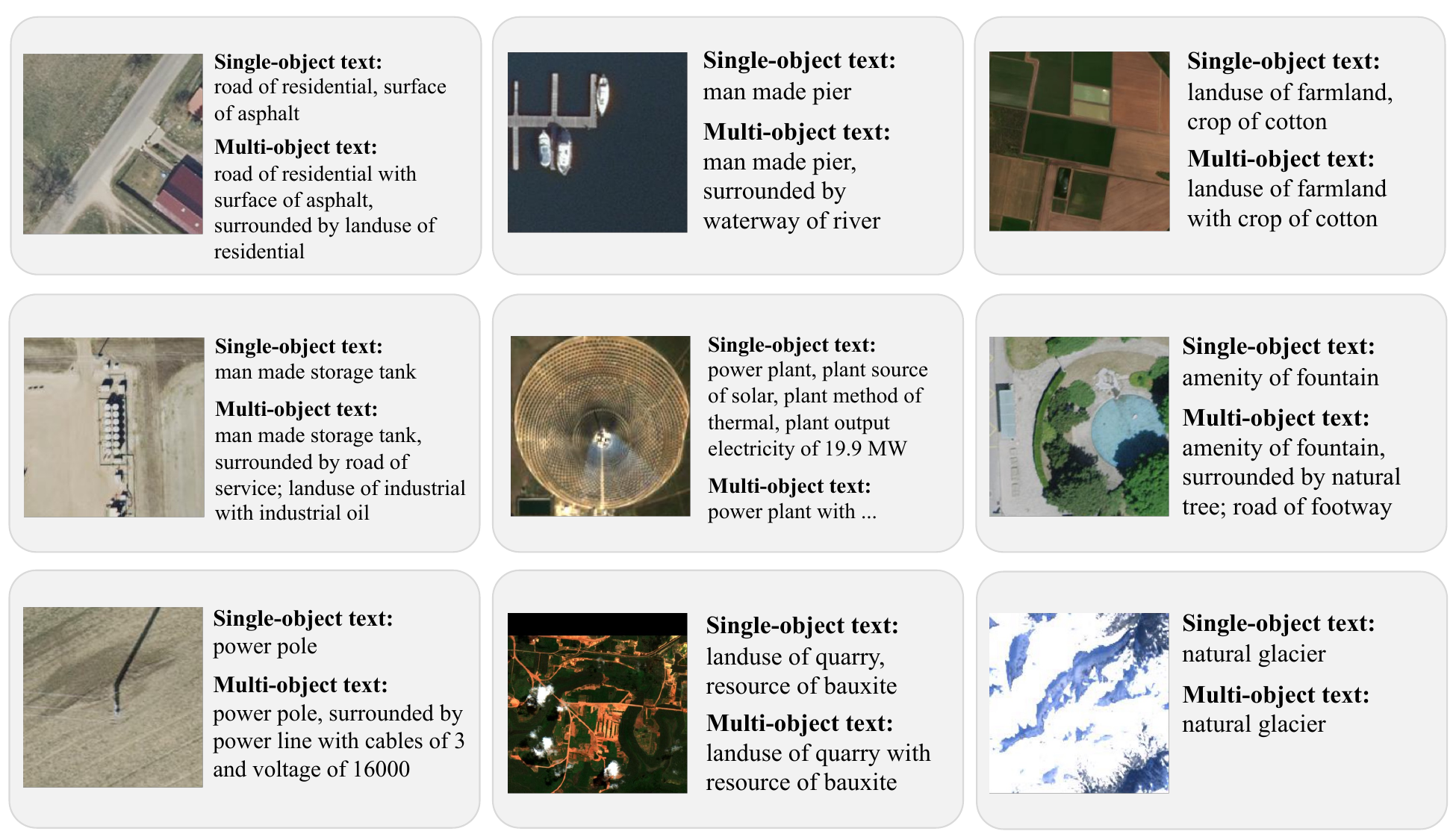}
    \caption{}
    \label{fig:examples}
  \end{subfigure}
  \hfill
  
  \caption{\textbf{(a)} The overview of the dataset construction approach. Remote sensing images of different sources are obtained from Google Earth Engine (GEE) platform, while semantic tags are obtained from OpenStreetMap (OSM). Images and tags are paired based on geo-coordinates.
      \textbf{(b)} Examples of image-caption pairs in the SkyScript dataset. Each image corresponds to a caption describing a single object and a caption describing multiple objects.}
  \label{fig:main_figure}
\end{figure*}

%% file: related_work.tex
\section{Related work}

\subsubsection{Vision-language model (VLM).}

Connecting images with their corresponding text descriptions has been shown to be an effective approach for learning visual representations. \cite{joulin2016learning,li2017learning,sariyildiz2020learning,desai2021virtex}. 
The development of CLIP and subsequent research further show that training on large-scale image-text datasets can yield VLMs that are generalizable to a wide variety of domains and robust to distribution shifts \cite{radford2021learning,jia2021scaling,li2022blip,yu2022coca,alayrac2022flamingo,chen2022pali}. 
In particular, CLIP aligns image and text representations with contrastive learning, enabling zero-shot transfer of the learned representations to various computer vision tasks in the open world \cite{radford2021learning}.
However, VLMs for remote sensing images are underexplored, constrained by the availability of large-scale remote sensing image-text dataset.


\subsubsection{Foundation models for remote sensing.}
Following the recent advancement of self-supervised pre-training in computer vision, research aiming at building remote sensing foundation models has explored two major directions. 
The first line of research aims to learn the representations of remote sensing images through establishing the similarity metrics between multiple images captured at different geo-locations or multiple views of the same object/location \cite{jean2019tile2vec,manas2021seasonal,ayush2021geography}. 
Another line of research attempts to train a vision model with the goal to reconstruct masked patches of the input image \cite{cong2022satmae,fuller2022satvit,fuller2022transfer,reed2022scale,mendieta2023gfm,cha2023billion}.
Both these two types of models involve imagery only, requiring further fine-tuning to adapt the model to different downstream tasks. 
The development of remote sensing VLMs integrating both imagery and text for zero-shot transfer is constrained by the availability of remote sensing image-text data.  
A contemporaneous work on remote-sensing-specialized CLIP \cite{liu2023remoteclip} relied on image-text pairs transformed from  existing remote sensing datasets, hence limited by the scale and semantic diversity of the existing data.

\subsubsection{Remote sensing dataset.} 

Owing to the domain knowledge needed for annotating remote sensing images, the construction of large and semantically diverse remote sensing datasets is particularly challenging. 
Only a few remote sensing datasets contain million-scale images (Figure \ref{fig:comparison}), but they cover only fixed sets of semantic classes no more than 150 \cite{yang2010bag,cheng2017remote,xia2017aid,christie2018functional,zhou2018patternnet,helber2019eurosat,sumbul2019bigearthnet,earthnets4eo,long2021creating,wang2021loveda,johnson2022opensentinelmap,bastani2023satlas,roberts2023satin}.
Existing remote sensing image-text datasets are particularly small \cite{Qu2016DeepSU,lu2017exploring,yuan2022exploring,Cheng2022NWPUCaptionsDA}, with a size ranging from several hundred to tens of thousand images. 
\citet{liu2023remoteclip} constructed a remote sensing image-text dataset containing 166K images by aggregating multiple existing remote sensing datasets and automatically assembling captions based on annotated bounding boxes or segmentation masks. 
However, the number of unique classes involved is bounded by the semantic diversity of existing remote sensing datasets.

%% file: dataset.tex
\section{Dataset}


\subsection{Data collection approach}

We construct the SkyScript dataset from the wild by linking large-scale yet unlabeled remote sensing image data with geo-tagged semantic information from OSM (Figure \ref{fig:image_text_register}). 
Here we introduce the data sources, the data selection approach, and how we derive image-text pairs.

\subsubsection{Source of images.} 
The data included in an open remote sensing image-text dataset ideally should have no licensing restrictions for research use. 
To this end, we acquire satellite and aerial images using the Google Earth Engine (GEE) platform which provides open access to large-scale remote sensing image collections from various sources allowing public sharing and redistribution. 
Specifically, Table \ref{tab:image_collections} lists the image collections used in SkyScript, forming a multi-source, multi-resolution image pool with ground sampling distance (GSD) ranging from 0.1 m/pixel to 30 m/pixel. 
For each image collection, we only consider RGB bands even if multispectral images are present. 
The inclusion of additional bands is left for future research.



\begin{table}[h]
\centering
\resizebox{1\columnwidth}{!}{
\begin{tabular}{lcc}
\toprule
Image collection       & GSD (m)  & Country  \\ 
\midrule
SWISSIMAGE 10 cm RGB imagery    & 0.1 & Switzerland          \\ 
Spain RGB orthophotos    & 0.1 & Spain          \\
Brandenburg (Germany) RGBN orthophotos    & 0.2 & Germany          \\
Finland RGB NLS orthophotos    & 0.5 & Finland          \\
National Agriculture Imagery Program    & 0.6-1 & U.S.          \\
Planet SkySat Public Ortho Imagery, RGB    & 0.8 & global          \\
Planet SkySat Public Ortho Imagery, MS   & 2 & global          \\
Harmonized Sentinel-2 MSI, Level-2A    & 10 & global          \\
Landsat 8 Collection 2 Tier 1 TOA Reflectance    & 30 & global          \\
Landsat 9 Collection 2 Tier 1 TOA Reflectance    & 30 & global          \\
\bottomrule
\end{tabular}}
\caption{Google Earth Engine (GEE) image collections used in the SkyScript dataset, together with their ground sampling distance (GSD) and country information. Abbreviations: MS: multispectral, MSI: multispectral instrument, TOA: top of atmosphere.}
\label{tab:image_collections}
\end{table}

\subsubsection{Source of semantics.} 
To enable the generalizability of VLM, the semantics covered in the dataset should ideally include not just a large variety of object categories, but also fine-grained subcategories and attributes. 
To bridge this gap for remote sensing images, we leverage the rich semantic information contained in OpenStreetMap (OSM), an open, crowdsourced geographic database. 
In OSM, each object on the map is described by one or more \textit{tags}. 
Each \textit{tag} consists of two free-form text fields, \textit{key} and \textit{value}. 
A \textit{key}, by definition, is used to describe a topic, a category, or a type of feature (e.g., ``surface''), while a \textit{value} describes the specific feature, attribute, or subcategory given the \textit{key} (e.g., ``asphalt''). 
Figure \ref{fig:image_text_register} shows more examples of tags.

Previously, the rich semantics in OSM have not been fully exploited in 
constructing remote sensing datasets for supervised learning, primarily due to the concern of its uncurated nature. 
However, the capability of noisy but semantically diverse image-text datasets based on web crawling have been demonstrated in contrastive image-text pre-training \cite{radford2021learning,jia2021scaling}. 
Our exploitation of uncurated yet rich semantics in OSM is based on this intuition.

\subsubsection{Connect images with appropriate semantics.} 
In OSM, some tags can be visually grounded in remote sensing images (e.g., ``waterway'': ``stream''), while others cannot at all (e.g., ``house number'': ``3''). 
Also, given an image of a certain GSD (e.g., 10 m), some tags that describe sufficiently large objects (e.g., ``natural'': ``coastline'') can be visually grounded at this image resolution while others cannot (e.g., ``power'': ``pole''). 

\begin{figure}[t]
\centering
\includegraphics[width=0.85\columnwidth]{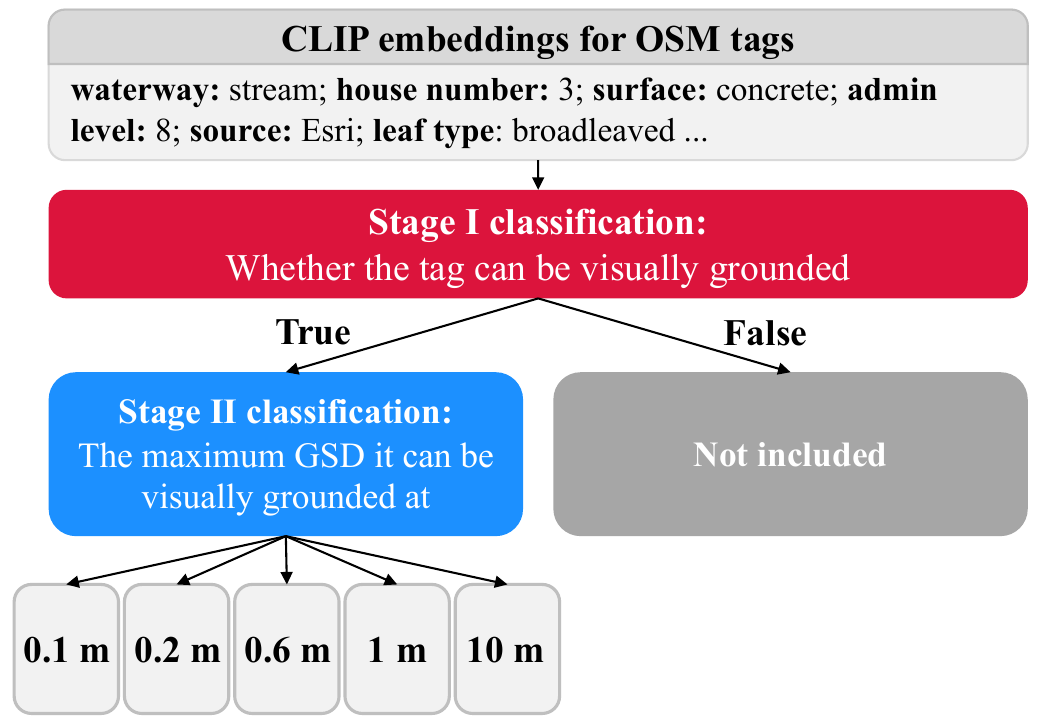}
\caption{Tag classification. Given the CLIP embedding of a tag, first determine whether it can be visually grounded in remote sensing images, and if yes, then determine the maximum ground sampling distance (GSD) at which it can be visually grounded. 
Manually-curated labels are provided to train classification models at both stages.
}
\label{fig:tag_classification}
\end{figure}

To determine which tags should be included to describe an image, we develop a two-stage tag classification approach with CLIP embeddings of tags (key + value) as inputs (Figure \ref{fig:tag_classification}). 
We use CLIP embeddings as they have already encoded visual information of tags after the image-text pre-training. 
In the first stage, a binary logistic regression model is used to predict whether the tag can be visually grounded at all. 
If a tag is predicted to be visually groundable, a second logistic regression model is further used to predict the maximum GSD (i.e., lowest resolution) at which the tag can be visually grounded. 
The prediction is one of the options: 0.1 m, 0.2 m, 0.6 m, 1 m, and 10 m. This is used to determine whether the tag should be included for describing an image given its GSD. 
More details about tag classification are provided in Appendix A.1.

\subsubsection{Data selection.} 
Data selection consists of two steps: object selection and image selection. 
To ensure global representativeness and semantic diversity, we perform a two-stage object selection based on OSM. 
The first stage is random object selection. 
Specifically, we randomly select 400K grids across the globe, each with 0.01$^{\circ}$ $\times$ 0.01$^{\circ}$ latitude/longitude intervals.  Then we query objects located in these grids from the OSM database. 
This enables the inclusion of a random, globally representative subset of objects with common tags (e.g., ``highway'': ``residential'', ``waterway'': ``river''). 
The second stage is targeted object selection. 
For rare tags that are not covered in the random object selection (e.g., ``archaeological site'': ``tumulus'', ``castle type'': ``palace''), we directly query all objects containing these tags from the OSM database. 
This enables the inclusion of rare semantics into our dataset as well. 
For both stages, OverPass API is used for querying the OSM database.

For image selection, we use a object-centered scheme to determine the image collection and tile boundary for each image tile. 
Specifically, for objects represented as points or polylines, we use the maximum allowable GSD predicted by our second-stage tag classification model to determine the suitable image collections of which the GSD is smaller than the maximum allowable GSD. 
Tile boundaries are initially determined by making the object point or a randomly selected node on the object polyline located at the image center. 
For objects represented as polygons, we combine the bounding box of the polygon, a range of desired image tile size, and GSD information of each image collection to determine the suitable image collections to use. 
The bounding box of the polygon is used as the image tile boundary. 
An object will be skipped if no suitable image collection can be found for it. 
To add variations of the object location in an image, we further alter the image tile boundary randomly so that the object deviates moderately from the image center. 
See Appendix A.2 for more details about image selection.

\begin{figure*}[t]
  \centering
  \begin{subfigure}{0.485\textwidth}
    \centering
   \includegraphics[width=\textwidth]{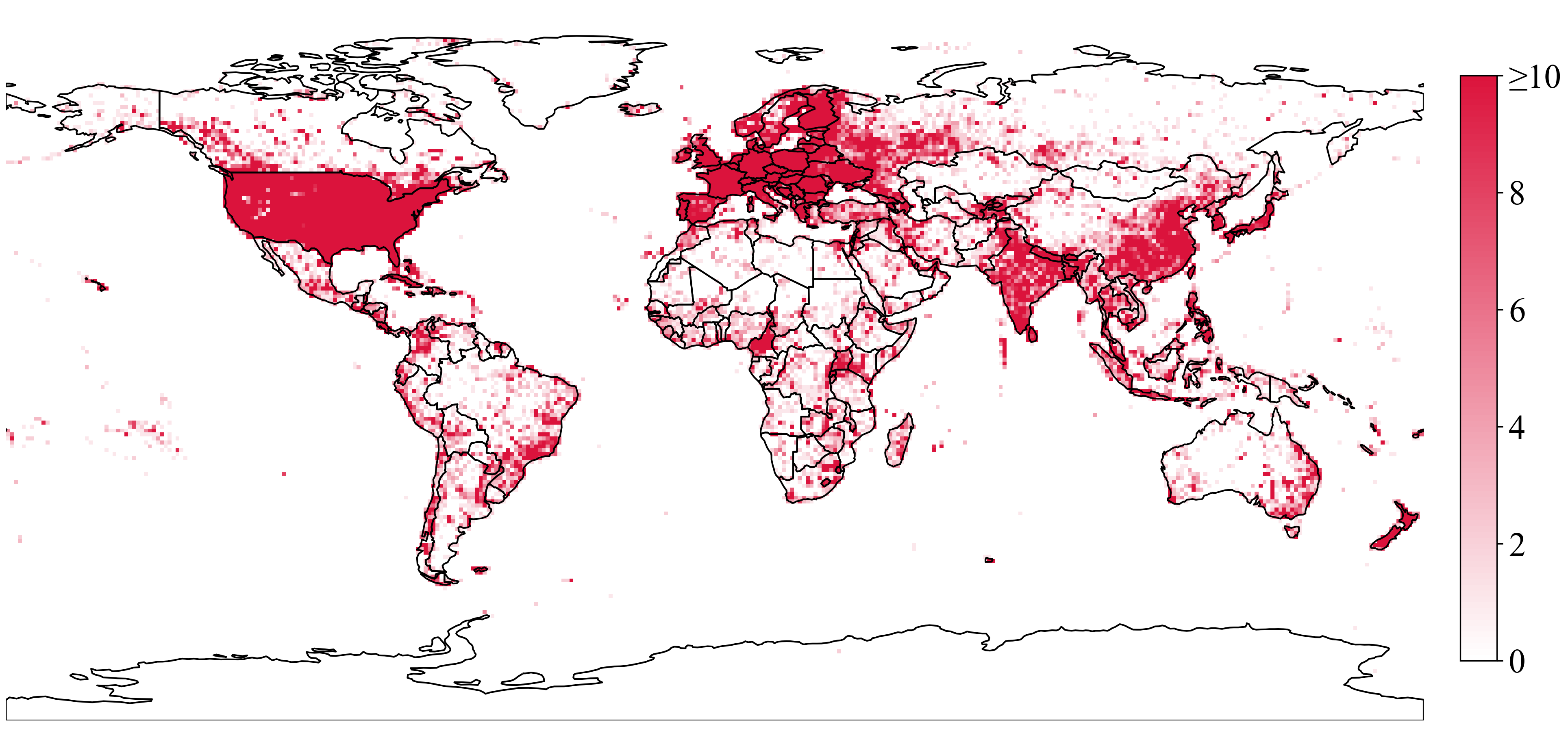}
    \caption{}
    \label{fig:coverage}
  \end{subfigure}%
  \hfill
  \begin{subfigure}{0.19\textwidth}
    \centering
   \includegraphics[width=\textwidth]{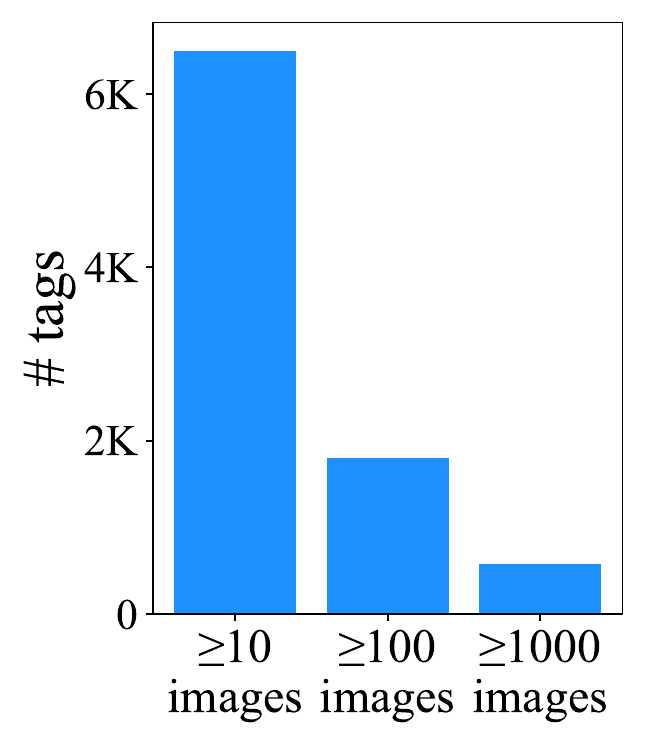}
    \caption{}
    \label{fig:tag_img_count}
  \end{subfigure}%
  \hfill
  \begin{subfigure}{0.32\textwidth}
    \centering
   \includegraphics[width=\textwidth]{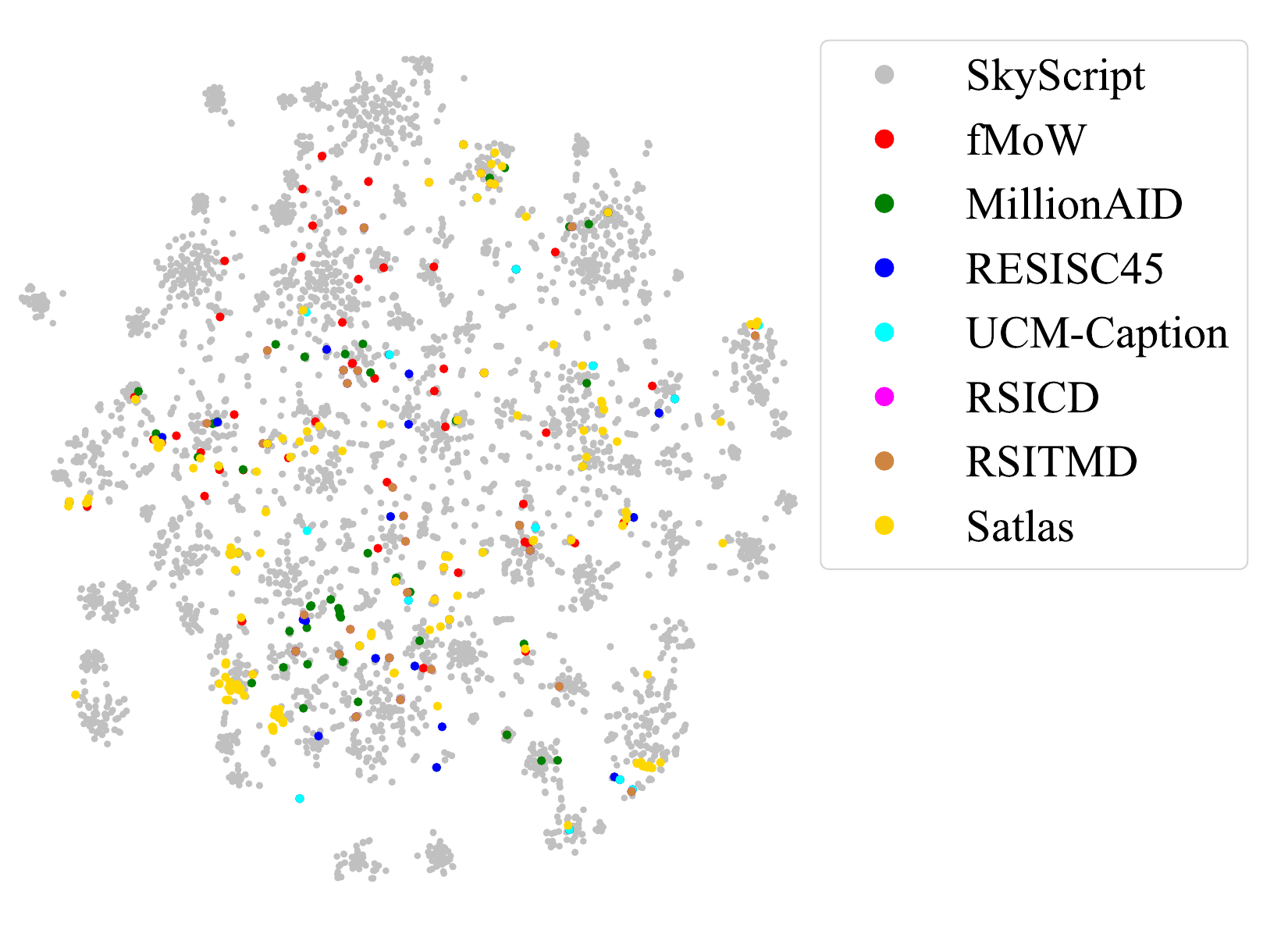}
    \caption{}
    \label{fig:tsne}
  \end{subfigure}
  \hfill
  \caption{\textbf{(a)} Geographic coverage of the SkyScript dataset, represented as the number of covered objects in each of 1$^{\circ}$ $\times$ 1$^{\circ}$ latitude/longitude grids. \textbf{(b)} Number of distinct tags with $\geq$10, $\geq$100, and $\geq$1,000 images included in SkyScript, respectively. \textbf{(c)} t-SNE visualization of semantic tag/label embeddings of different remote sensing datasets. }
  %
  \label{fig:datset_analysis}
\end{figure*}

\subsubsection{Assembling caption.}
Each object can have multiple tags, and each tag consists of a key and a value. 
We convert tags into a caption by first connecting the key and the value with a connecting word (e.g., ``of'', ``is'') and then connecting multiple tags with comma or ``and''. 
Appendix A.3 provides details about the rules used for assembling caption.

For each image tile, we generate two captions. 
The first caption describes only the object that is used for determining the image tile boundary. 
We also obtain other OSM objects contained in the image tile by using geospatial overlay and then assemble a second caption describing multiple objects in the image tile.
For example, if a object with a tag \{``power'': ``pole''\} is surrounded by another object with tags \{``power'': ``minor line'', ``cables'': ``3'', ``voltage'': ``16000''\}. Then the two captions are: \\

Single-object caption: \textit{power pole}

Multi-object caption:  \textit{power pole, surrounded by power minor line with cables of 3 and voltage of 16000}

\subsubsection{Filtering out uncorrelated image-text pairs.}
To reduce noisy information in this wild, uncurated dataset, we filter out the image-text pairs where the image and its corresponding caption are not sufficiently relevant. 
Specifically, after a caption is assembled from tags, we apply OpenAI's ViT-L/14 CLIP model to obtain the image and caption embeddings for each image-caption pair and calculate their cosine similarity. 
Low-similarity pairs indicate noisy samples (e.g., the object described by the caption is obscured by trees). 
We only keep image-caption pairs with a similarity value among the top. 
We experiment with top 20\%, 30\%, and 50\% and the results are discussed in the next section.

\subsection{Dataset analysis}

\subsubsection{Dataset overview.}
Using the data collection approach described above, we obtain 5.2 million unfiltered remote sensing image-text pairs covering 44K distinct tags. 
Figure \ref{fig:examples} show examples of image-text pairs. 
By keeping only image-text pairs with a similarity value among the top 50\%, we obtain 2.6 million image-text pairs covering 29K distinct tags. 
These tags form 100K distinct single-object captions and 1.2 million distinct multi-object captions. 
GSD of images range from 0.1 m to 30 m. 
We randomly sample 1,000 image-tag pairs and manually check the tag accuracy, which is 96.1\%.

30,000 image-text pairs are set aside for testing cross-modal retrieval performance. 
This test set is denoted as ``SkyScript-retrieval''.
We also collect additional samples to form a classification dataset containing 70 classes for validating classification performance. 
Objects in this set are not covered by the main dataset. 
This auxiliary dataset, denoted as ``SkyScript-classification'', is detailed in Appendix A.4.

\subsubsection{Geographic coverage.} 
As Figure \ref{fig:coverage} shows, SkyScript has a good geographic coverage for all continents except Antarctica. 
The U.S. and Europe have a particularly high volume of objects covered, as high-resolution images ($<$1 m) required for visually grounding small objects are concentrated in these two regions. 
In the rest of the world, regions with denser population tend to have more objects covered (e.g., east China, India). 
This reflects higher object density (e.g., building) and probably more complete OSM annotations in these densely-populated areas. 

\subsubsection{Semantic diversity.} 
SkyScript is semantically diverse. 
As Figure \ref{fig:tag_img_count} shows, 580 tags have $\geq$1,000 images, and more than 1,800 tags have $\geq$100 images included in the dataset. 
By using t-SNE to project the CLIP embeddings of tags covered in SkyScript as well as the semantics classes of other datasets into the 2D space (Figure \ref{fig:tsne}), we find that the semantics covered in the SkyScript dataset can be viewed as a superset of those covered in previous datasets. 
As demonstrated in Figure \ref{fig:examples}, SkyScript includes not only broad category information, but also fine-grained information about object attributes (e.g., crop type of farmland, road surface, power plant source and capacity).

\subsubsection{Comparison with the remote sensing subset in LAION.} LAION-2B is a huge English text-image dataset containing 2.3B images and their English captions obtained by web crawling \cite{schuhmann2022laion}. 
We obtain a remote sensing subset of it by applying a binary classification model to determine whether an image in LAION-2B is a remote sensing image (see details in Appendix A.5). 
This subset, denoted as LAION-RS, contains 726K remote sensing image-text pairs—only 0.03\% of all samples. 
This shows that web crawling cannot efficiently collect remote sensing image-text pairs at scale. 
We compare SkyScript with LAION-RS by evaluating them in downstream tasks.

\begin{table*}[h]
\centering
\resizebox{\linewidth}{!}{
\begin{tabular}{llcccccccccccc}
\toprule
\multirow{2}{*}{ViT} & \multirow{2}{*}{Model} & \multicolumn{9}{c}{Scene classification} & \multicolumn{3}{c}{Fine-grained classification} \\
\cmidrule(lr){3-3} \cmidrule(lr){4-11} \cmidrule(lr){12-14}
& & SkyScript & AID & EuroSAT & fMoW & Million-AID & PatternNet & RESISC & RSI-CB & Avg. & Roof shape & Smoothness & Surface \\
\midrule
\multirow{5}{*}{B-32} & CLIP-original & 40.16 & 69.55 & 32.11 & 17.62 & 57.27 & 64.09 & 65.71 & 41.26 & 49.66 & 31.50 & 26.80 & 61.36 \\
& Human-curated captions & 40.03 & 71.05 & 33.85 & 18.02 & 57.48 & 66.56 & 66.04 & 42.73 & 50.82 & 28.50 & 27.80 & 60.91 \\
& RemoteCLIP  & 27.06 & \textbf{87.05} & 30.74 & 11.13 & 46.26 & 56.05 & 67.88 & 44.55 & 49.09 & 30.50 & 21.00 & 43.86 \\
& CLIP-laion-RS & 40.77 & 69.55 & 37.63 & 19.16 & 56.59 & 64.79 & 64.63 & 41.79 & 50.59 & 28.83 & 27.60 & 62.27 \\
& SkyCLIP-50 & 52.98 & 70.90 & 33.30 & 19.24 & 62.69 & 72.18 & 66.67 & 46.20 & 53.02 & 26.00 & \textbf{38.00} & \textbf{67.73} \\
\midrule
\multirow{7}{*}{L-14} & CLIP-original & 55.06 & 69.25 & 41.89 & 26.19 & 57.88 & 71.39 & 66.70 & 43.02 & 53.76 & 37.50 & 25.40 & 42.73 \\
& Human-curated captions & 56.09 & 72.95 & 41.96 & 26.33 & 58.47 & 74.86 & 68.70 & 44.60 & 55.41 & 37.00 & 26.60 & 40.00 \\
& RemoteCLIP  & 34.40 & 70.85 & 27.81 & 16.77 & 47.20 & 61.91 & \textbf{74.31} & \textbf{50.79} & 49.95 & 34.33 & 34.20 & 55.45 \\
& CLIP-laion-RS & 58.81 & 71.70 & \textbf{54.30} & 27.21 & 60.77 & 72.68 & 71.21 & 48.21 & 57.87 & 40.50 & 37.60 & 53.41 \\
& SkyCLIP-20 & 67.94 & 71.95 & 53.63 & \textbf{28.04} & 65.68 & 78.62 & 70.70 & 50.03 & 59.81 & 44.83 & 26.80 & 61.36 \\
& SkyCLIP-30 & 69.08 & 72.15 & 52.44 & 27.77 & 66.40 & 79.67 & 70.77 & 50.19 & 59.91 & 46.17 & 30.80 & 64.32 \\
& SkyCLIP-50 & \textbf{70.89} & 71.70 & 51.33 & 27.12 & \textbf{67.45} & \textbf{80.88} & 70.94 & 50.09 & \textbf{59.93} & \textbf{46.83} & 35.80 & 67.50 \\
\bottomrule
\end{tabular}}
\caption{Top-1 accuracy (\%) for zero-shot scene classification and fine-grained classification.}
\label{tab:cls_results}
\end{table*}

\subsection{Applications and limitations}

SkyScript can be used for developing models for a variety of tasks in remote sensing, such as open-vocabulary classification, cross-modal retrieval, image captioning, and text-to-image generation.  
It has potential values in a broad range of applications for sustainable development, such as monitoring the conditions of infrastructures (e.g., road, bridge), identifying illegal mining, tracking land use, and mapping distributed renewable energy resources.
SkyScript alone can be used to develop domain-specialized VLMs for remote sensing by using continual pre-training. 
It can also be combined with other image-text datasets together to pre-train general-purpose VLMs from scratch. 

SkyScript may have inherent bias. 
First, as currently we only consider remote sensing imagery without licensing restriction, high-resolution images are mainly limited to the U.S. and Europe, making the rest of the world underrepresented in the dataset. 
Second, OSM annotations are less complete in developing countries, which, again, makes the samples in these regions less abundant. 
Potential mitigation approaches include partnership with Earth observation companies or government agencies to obtain high-resolution images with wider coverage, as well as taking both models and human annotations in the loop to annotate more objects in underrepresented countries and regions. 
Moreover, we only use a simple rule-based approach to automatically assemble captions from tags. 
Using large language models (LLMs) to generate more natural and meaningful captions from tags warrants future exploration.

%% file: experiments.tex
\section{Experiments}

\begin{table*}[htb]
    \begin{minipage}{0.67\linewidth}
    \resizebox{\linewidth}{!}{
        \begin{tabular}{lcccccccc}
        \toprule
        \multirow{2}{*}{Model} & \multicolumn{2}{c}{SkyScript-retrieval} & \multicolumn{2}{c}{RSICD} & \multicolumn{2}{c}{RSITMD} & \multicolumn{2}{c}{UCM-Captions} \\
        \cmidrule(lr){2-3} \cmidrule(lr){4-5} \cmidrule(lr){6-7} \cmidrule(lr){8-9}
        & img2txt & txt2img & img2txt & txt2img & img2txt & txt2img & img2txt & txt2img \\
        \midrule
        \multicolumn{9}{l}{Supervised cross-modal retrieval models} \\
        \midrule
        AMFMN & {-} & {-} & (14.62) & (18.21) & (25.74) & (33.69) & (43.65) & (48.51) \\
        LW-MCR & {-} & {-} & (12.68) & (18.50) & (24.70) & (32.45) & (43.02) & (47.68) \\
        GaLR & {-} & {-} & (19.16) & (18.77) & (29.65) & (33.17) & - & - \\
        \midrule
        \multicolumn{9}{l}{Vision-language models} \\
        \midrule
        CLIP-original & 2.97 & 1.95 & 19.67 & 13.84 & 27.51 & 24.10 & 68.41 & 56.76 \\
        Human-curated captions & 3.28 & 2.18 & (20.56) & (16.37) & 27.88 & 28.47 & (70.95) & (59.59) \\ 
        CLIP-laion-RS & 3.85 & 2.81 & 22.66 & 18.52 & 30.24 & 29.67 & 69.68 & 57.56 \\ 
        RemoteCLIP & 5.08 & 2.81 & (36.32) & (33.20) & (43.95) & (44.94) & (79.05) & (74.98) \\
        
        SkyCLIP-30 & (8.53) & (7.73) & 23.70 & 19.97 & 30.75 & 30.58 & 72.22 & 59.33 \\
        \bottomrule
        \end{tabular}}
        \caption{Mean recall (\%) for cross-modal retrieval. 
        SkyCLIP-30 is trained on SkyScript (top 30\% samples in terms of  pairwise similarity) using multi-object captions. If a dataset is involved in training (``in-domain''), then it is bracketed with ``()''.
        }
      \label{tab:retrieval_results}
    \end{minipage}
    \hspace{0.01\linewidth} 
    \hfill
    \begin{minipage}{0.29\linewidth} 
        \centering
        \includegraphics[width=\linewidth]{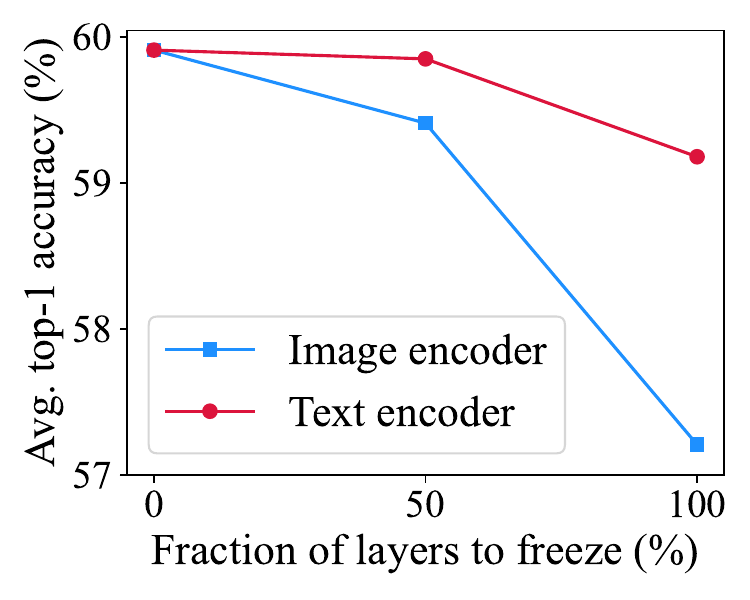} 
        \captionof{figure}{Effect of layer freezing on zero-shot classification accuracy (averaged over benchmark datasets).}
        \label{fig:layer_freezing}
    \end{minipage}
\end{table*}

We demonstrate the value of the SkyScript dataset by using it to develop a CLIP model for remote sensing through continual pre-training. 
This model is further evaluated in zero-shot classification and cross-modal retrieval, showing substantive performance gains compared with the VLMs pre-trained on web-based image-text datasets. 

\subsection{CLIP continual pre-training} 

We perform continual pre-training on SkyScript to obtain a remote-sensing-specialized CLIP model, denoted as SkyCLIP. 
Specifically, we initialize the CLIP model with the weights that were pre-trained on web-based image-text data, and then further train it on the SkyScript dataset using image-text contrastive learning \cite{radford2021learning}. 
We consider two ViT versions for CLIP: CLIP/ViT-B-32 and CLIP/ViT-L-14. 
CLIP/ViT-B-32 tokenizes images by patches of 32 $\times$ 32 pixels and has 12 transformer layers. 
By contrast, CLIP/ViT-L-14 uses a larger ViT that tokenizes images by patches of 14 $\times$ 14 pixels and has 24 transformer layers. 
For CLIP/ViT-B-32, we use the LAION-2B model weights to initialize the model, while for CLIP/ViT-L-14, we use the OpenAI model weights to initialize the model, both based on the performances of initial model weights. 
The continual pre-training is conducted on 4 NVIDIA A100 GPUs with a batch size of 512 and total epochs of 20.

\subsection{Zero-shot scene classification}

\subsubsection{Benchmark datasets.} 
We evaluate the zero-shot scene classification performance of SkyCLIP and other baseline models on seven common and comparatively large benchmark datasets: AID \cite{xia2017aid}, EuroSAT \cite{helber2019eurosat}, fMoW \cite{christie2018functional}, Million-AID \cite{long2021creating}, PatternNet \cite{zhou2018patternnet}, NWPU-RESISC45 \cite{cheng2017remote}, and RSI-CB256 \cite{Li2017RSICBAL}. 
Appendix A.6 provides more details about benchmark datasets.
As a reference, we also evaluate the performance on the SkyScript classification dataset. 

\subsubsection{Definition of zero-shot transfer.} 
Following the use of the term ``zero-shot'' in \citep{radford2021learning}, rather than measuring the model generalizability on unseen object categories, here we evaluate the ``zero-shot'' transfer capability on unseen datasets. 
Although we cannot eliminate the possibility of SkyScript overlapping with the benchmark datasets in terms of geo-locations, they are unlikely to overlap in terms of images. 
This is because images in SkyScript are obtained from open image collections from GEE, while those in benchmark datasets are usually from commercial satellite images captured by different cameras, hence rendering different images even at the same geo-location.

\subsubsection{Results.}
We use the original CLIP models as baselines, denoted as ``CLIP-original''. 
We also use continual pre-training to train a CLIP model on the LAION-RS dataset, denoted as ``CLIP-laion-RS''. 
Additionally, we include RemoteCLIP \cite{liu2023remoteclip} and a dataset aggregating existing human-curated captioning datasets (NWPU-Captions, UCM-Captions, Sydney-Captions, RSICD) for comparison.
This aggregated dataset is denoted as ``human-curated captions''.
Models trained on SkyScript are differentiated by the fraction of samples included (top 20\%, 30\%, or 50\%) based on the descending order of CLIP similarity of image-text pairs, denoted as ``SkyCLIP-20'', ``SkyCLIP-30'', and ``SkyCLIP-50'', separately.

Table \ref{tab:cls_results} shows the top-1 accuracies on different scene classification datasets. 
We find that SkyCLIP not only significantly improves the performance on the in-domain SkyScript classification dataset, 
but also consistently generalizes better to unseen benchmark datasets than the baseline models. 
For SkyCLIP-50 with the ViT-L-14 backbone, the average top-1 accuracy is 59.93\%, with a 6.17\% gain compared with the original CLIP/ViT-L-14, and a 2.06\% gain compared with the CLIP model trained on the remote sensing subset of LAION-2B (CLIP-laion-RS).
This suggests that SkyScript can yield a VLM with better zero-shot transfer capability on unseen remote sensing datasets than the general VLM. 
Also, simply taking the remote sensing subset of web-scale data for continual pre-training cannot reach the same performance as training on SkyScript.
Moreover, SkyCLIP-50 outperforms RemoteCLIP (+9.98\% average accuracy) and the model developed with human-curated captions (+4.52\% average accuracy), indicating the importance of semantic diversity of image-text datasets in developing VLMs for remote sensing images.

As Table \ref{tab:cls_results} shows, SkyCLIP-50 outperforms SkyCLIP-20 and SkyCLIP-30 on some benchmark datasets (Million-AID, PatternNet, RESISC45) but not on others. 
This reflects the trade-off between increasing the number of image-text pairs and ensuring their high quality as measured by their predicted pairwise similarity.

\subsubsection{The effect of layer freezing.} 

Since our models are trained with the continual pre-training setting, we also investigate whether training only a fraction of downstream layers while freezing the rest can yield better performance. 
As Figure \ref{fig:layer_freezing} shows, freezing more layers in either the image encoder or text encoder of CLIP degrades the performance. 
Compared with freezing the text encoder, freezing the image encoder yields worse accuracy.
This implies that both image and text representations need to be adjusted to adapt from the general domain to the remote sensing domain, and the adjustment of image representations is more critical.

\subsection{Zero-shot fine-grained classification}
Compared with previous remote sensing datasets, the rich semantics covered in SkyScript can further facilitate the learning of fine-grained subcategories or attributes. 
Here we also demonstrate the fine-grained classification ability of the CLIP trained with SkyScript by testing its zero-shot performance on classifying three different attributes: roof shape, road smoothness, and road surface materials. 

\subsubsection{Test data.} We construct a test set for evaluating the classification of each of the three attributes. 
Roof shapes include 6 classes: flat, hipped, gabled, dome, pyramidal, and round. 
Road smoothness includes 5 classes: excellent, good, intermediate, bad, and very bad. 
Road surface includes 5 classes: asphalt, concrete, grass, gravel, and sand. 
Each class contains 60 to 100 images. 
To abide by the zero-shot principle, we use Google and Bing Maps API instead of GEE to collect these images and do not consider the objects that have already been included in SkyScript. 
This ensure that images and objects in these fine-grained classification test sets are not seen during training. 
Appendix A.7 provides more details about the test sets.

\subsubsection{Results.} The rightmost panel of Table \ref{tab:cls_results} shows the top-1 accuracies of SkyCLIP and other models on the fine-grained classification test sets. 
For roof shape classification, SkyCLIP with the ViT-L-14 image encoder achieves the highest top-1 accuracy (46.83\%), while for road smoothness and surface material classification, SkyCLIP with the ViT-B-32 image encoder, though smaller, achieves the highest top-1 accuracies (38.00\% and 67.73\%). 
For both ViT-B-32 and ViT-L-14, SkyCLIP outperforms its CLIP counterpart by a significant margin (+6.37\% to +24.77\%), except a performance decrease of -5.5\% for ViT-B-32 on roof shape classification. 
It is noteworthy that the performance gain of SkyCLIP in zero-shot fine-grained classification is more substantial than that on scene classification.
This suggests the rich semantics covered in SkyScript can benefit those remote sensing applications without abundant labeled data, such as monitoring the conditions of civil infrastructure.

\subsection{Cross-modal retrieval}
We further demonstrate the value of SkyScript on cross-modal retrieval. 
In this task, given the query in image/text, the model is intended to find the best match of text/image in the dataset.
This is done by calculating the cosine similarity between the query image/text embedding and the embedding of each text/image candidates.

\subsubsection{Benchmark datasets.} 
We evaluate the cross-retrieval performance on the test set of SkyScript (``in-domain'') containing 30,000 image-text pairs, together with three benchmark image-text datasets for remote sensing (``zero-shot''): UCM-Captions \cite{Qu2016DeepSU}, RISCD \cite{lu2017exploring}, and RSITMD \cite{yuan2022exploring}. 

\subsubsection{Results.}
In addition to the comparison with CLIP-based models,
We also compare SkyCLIP with three recently-developed remote sensing cross-modal retrieval models: AMFMN \cite{yuan2022exploring}, LW-MCR \cite{Yuan2022ALM}, and GaLR \cite{Yuan2022RemoteSC}. 
Benchmark datasets were seen during the training of these three models.

The mean recall, defined as the average of recall@1, recall@5, and recall@10, is used to measure the overall performance for both image-to-text and text-to-image retrieval. 
As Table \ref{tab:retrieval_results} shows, SkyCLIP can steadily achieve better retrieval performance than CLIP and CLIP-laion-RS on three benchmark datasets that have never been seen during training (+2.57\% to +6.48\%). 
Its mean recalls on the benchmark datasets are comparable or even higher than the three supervised models (AMFMN, LW-MCR, GaLR) that have already seen the benchmark datasets during training. 
Notably, on UCM-Captions dataset, SkyCLIP outperforms the supervised models by $>$10\% for text-to-image retrieval and $>$25\% for image-to-text retrieval.
This demonstrates the versatility of the visual and text representations learned from SkyScript, which can be transferred to unseen cross-modal remote sensing tasks in a zero-shot setting.

%% file: conclusion.tex
\section{Conclusion}

We present SkyScript, a large and semantically diverse image-text dataset for remote sensing images. 
We demonstrate its value by using it to derive a remote-sensing-specialized CLIP model outperforming baseline models across three downstream tasks in the zero-shot setting: scene classification, fine-grained classification, and cross-modal retrieval.
The limitation in its geographic representativeness has also been discussed. 
Future work can explore its usage in other remote sensing tasks such as image captioning and text-to-image generation. 

%% file: ackowledgements.tex
\section{Acknowledgements}

This work is in part supported by the Stanford Human-Centered AI (HAI) Fellowship, Stanford Data Science Scholarship, U.S. Department of Energy DE-EE0009359, and Stanford Precourt Institute for Energy.

%% file: appendix.tex
\subsection{A.1. Tag classification}

We develop a two-stage tag classification approach to determine whether a tag should be included to describe an image. 
Each stage is a logistic regression model taking the CLIP embedding of a tag as input. 
The embedding is outputted by the text encoder of the pre-trained CLIP/ViT-L-14. 

The first stage is a binary logistic regression model that predicts whether a tag can be visually grounded in remote sensing images at all. 
2,000 tags are labeled to develop the first-stage model. 
We use 5-fold cross validation to select the best model hyperparameters. 
The best cross-validation F1 score (harmonic mean of precision and recall) is 0.88.

The second stage is a multi-class logistic regression model that further determines the maximum ground sampling distance (GSD) a tag can be visually grounded. 
Output options include: 0.1 m, 0.2 m, 0.6 m, 1 m, and 10 m. These cut-off GSDs are selected based on the GSD of GEE image collections used for dataset construction (listed in the main paper Table 1). 
514 tags are labeled to develop the second-stage model. 
To reduce overfitting, PCA is used to reduce the dimension of CLIP embedding before it is fed into the logistic regression model.
We use 5-fold cross validation to select the best model hyperparameters. 
The best cross-validation accuracy is 0.53. 
The second-stage model is applied to a tag only if the tag is predicted to be visually groundable at the first stage.

\subsection{A.2. Image selection}

\subsubsection{A.2.1. Image collection selection.}
If there are multiple GEE image collections available at the geo-location of an OSM object, then we perform image collection selection. 
The first step is to identify whether an image collection is suitable for visually grounding the object. 
Specifically, for OSM objects represented as points or polylines, an image collection is considered as suitable if its GSD is smaller than the maximum allowable GSD predicted by the second-stage tag classification model.
For an OSM object represented as a polygon, we use the bounding box of the polygon as the image boundary, and an image collection is considered as suitable if it can yield an image for this object with both the height and width  between 75 and 1,000 pixels. 
As the second step, if there are multiple suitable image collections available for an object, we randomly select one of them. 

\subsubsection{A.2.2. Image boundary selection.} 
Image boundary selection is to determine the geo-coordinates of the four corners of an image. 
We first determine the basic image boundary. 
Specifically, for an OSM object represented as a point, we set the image boundary so that the image has a size of 224 $\times$ 224 centering at this point. 
Similarly, for an OSM object represented as a polyline, we randomly select a node on the polyline and apply the same approach to this node so that it is at the image center. 
For an OSM object represented as a polygon, we directly use the bounding box of the polygon as the basic image boundary.

Then we randomly alter basic image boundaries to add more variation to the images included in the dataset. 
For an object represented as a point or a polyline, we make the point (or the selected node of the polyline) moderately deviate from the image center but still located within the middle 1/3 part of the image. 
Image height and width are re-selected randomly within a range from 168 to 300 pixels. 
For an object represented as a polygon, image height and width are also randomly re-selected, but within a range from 150 to 1,500 pixels and with an aspect ratio between 0.5 to 2. 

Finally, we download images from GEE by providing the image collection and image boundary information.

\subsection{A.3. Assembling caption}
Each object can have one or more tags, which are assembled into a caption following a set of designed rules. 
It consists of three steps:

\begin{itemize}
    \item \textbf{Rename key or value.} We change some key or value names in OSM as they are not the most commonly-used text descriptions for a specific type of objects. 
    For example, ``highway'' is renamed as ``road'' if it is used to describe roads other than motorway, trunk road, or primary road (e.g., residential road). 
    ``aeroway'' is renamed as ``airport''. 
    ``lit'' is renamed as ``light''. 
    ``leisure'' is renamed as ``leisure land''.
    
    \item \textbf{Connect key and value.} 
    If the key is an adjective (e.g., ``natural''), then simply use a space to connect key and value (e.g., \{``natural'', ``water''\} $\rightarrow$ ``natural water''). 
    If the key is an attribute name (e.g., ``smoothness'', ``visibility''), then use ``is'' to connect the key and value (e.g., \{``smoothness'', ``good''\} $\rightarrow$ ``smoothness is good''). 
    If the value is ``construction'' which means an object is under construction, then use ``under'' as the connecting word (e.g., \{``building'', ``construction''\} $\rightarrow$ ``building under construction''). 
    In other cases, use ``of'' as the default connecting word (e.g.,  \{``lanes'', ``2''\} $\rightarrow$ ``lanes of 2'').

    \item \textbf{Connect multiple tags.} 
    For single-object captions, we connect the phrases of tags with commas. 
    For multi-object captions, we connect the phrases of tags of each object with ``with'' followed by ``and''.
    The assembled description of the center object is connected with the descriptions of surrounding objects using ``surrounded by''.  
    
\end{itemize}

For example, if a object with a tag \{``power'': ``pole''\} is surrounded by another objects with tags \{``power'': ``minor line'', ``cables'': ``3'', ``voltage'': ``16000''\}. Then the two captions are: \\

Single-object caption: \textit{power pole}

Multi-object caption:  \textit{power pole, surrounded by power minor line with cables of 3 and voltage of 16000}  \\

The full algorithm for assembling captions can be found in the Supplementary Code. 
Simple caption assembly without compositional structure information and without better alignment with human language can make VLMs trained on the dataset behave like bag-of-words \cite{yuksekgonul2022and}. 
Generating more natural and complicated captions has the potential to further improve the performance of VLMs, and it is part of our future work. 
We will also release the raw tags corresponding to each image in addition to the captions to support the development of better approaches to generate captions from tags.

\subsection{A.4. SkyScript classification dataset}

We collect additional images to form an ``in-domain'' test set for scene classification, denoted as ``SkyScript-classification''. 
Although these images are also collected from GEE, objects in this dataset are not covered by the main SkyScript dataset. 

SkyScript-classification has 70 classes. They are: airport apron, airport hangar, airport runway, airport terminal, amusement park, apartment, aquaculture land, archaeological site, barn, baseball field, basketball court, beach, border checkpoint, bridge, cemetery, church, circular farmland, commercial land, construction land, crosswalk, dam, detached house, educational institution, factory, fire station, forest, freeway, gas station, golf course, greenhouse, ground transportation station, harbour, helipad, hospital, glacier, island, lake, lighthouse, meadow, warehouse, mine, mountain, office building, oil well, orchard, parking space, playground, pond, prison, quarry, race track, railway, railway station, residential land, resort, river, rock land, roundabout, solar power plant, square, stadium, storage tank, substation, swimming pool, tennis court, tower, turning circle, wastewater treatment plant, wetland, wind farm.

Every class has 100 images except a few rare classes including archaeological site, fire station, lighthouse, resort, and solar power plant. 
These images and their class labels are manually validated by human, and will also be made public upon publication.

\begin{table}[t]
\centering
\resizebox{0.85\columnwidth}{!}{
\begin{tabular}{lccc}
\toprule
Dataset name  & Partition  & \# classes  & \# images  \\ 
\midrule
AID         & test  & 30   & 2,000        \\ 
EuroSAT     & test  & 10   & 2,700       \\
fMoW        & val   & 62   & 106,081       \\
Million-AID & train & 51   & 10,000       \\
PatternNet  & train & 38   & 30,400       \\
RESISC45    & train & 45   & 31,500       \\
RSI-CB256   & train & 35   & 24,747       \\

\bottomrule
\end{tabular}}
\caption{Benchmark datasets used for testing scene classification.}
\label{tab:benchmark_datasets}
\end{table}

\subsection{A.5. Remote sensing subset of LAION-2B}

We extract a remote sensing image subset from LAION-2B, a large-scale English text-image dataset containing 2.3B web images to test whether it can yield a better VLM in the remote sensing domain. 
Specifically, we take the image embeddings outputted by the image encoder of the original CLIP/ViT-L-14 as input features and use them to train a logistic regression model to predict whether a given image is remote sensing image or not. 
We curate a dataset with 2,000 remote sensing images and 50,000 non-remote-sensing images from LAION-2B. 
We use 90\% of them for training and 10\% for testing. 
The accuracy on the test set is 99.96\%, indicating the effectiveness of CLIP embedding in distinguishing between remote sensing images and non-remote sensing images. 
This logistic regression model is further applied to all image embeddings in LAION-2B, resulting a remote sensing subset containing 726K image-text pairs, denoted as LAION-RS.

\subsection{A.6. Benchmark datasets for scene classification}

Table \ref{tab:benchmark_datasets} lists the dataset partition, number of semantic classes, and number of images in each benchmark dataset used for testing scene classification. 
For some benchmark datasets, we use the training set partition because of the unavailability of images or labels for their test set partitions, but none of the partitions (train, val, and test) of each benchmark dataset is seen during the continual pre-training of CLIP.

\subsection{A.7. Fine-grained classification test set}

\begin{figure*}[t]
\centering
\includegraphics[width=0.7\linewidth]{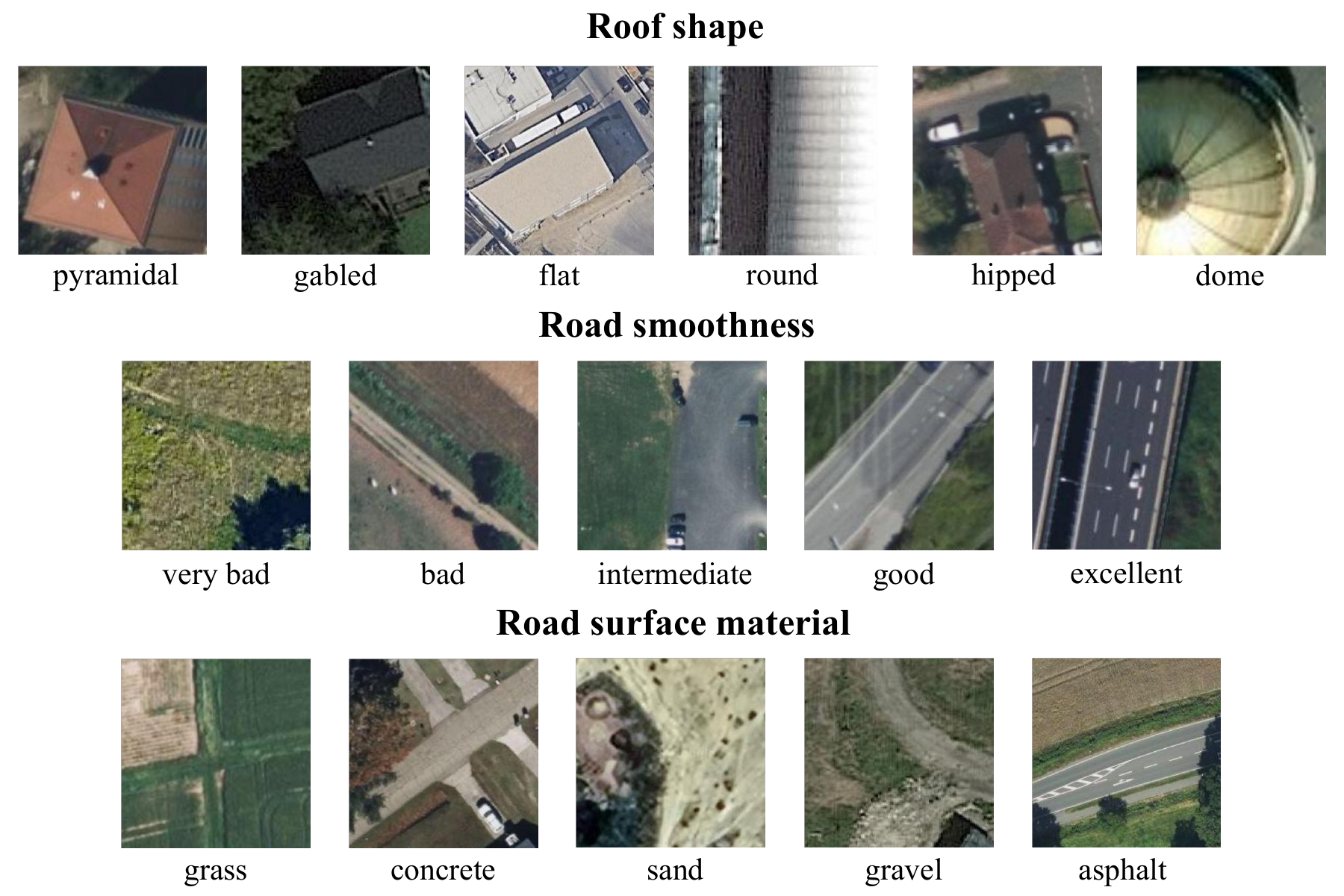}
\caption{Examples of images in each class of the three fine-grained classification tasks: roof shape classification, road smoothness classification, and road surface material classification.
}
\label{fig:fine_grained}
\end{figure*}

We curate a separate test set for testing zero-shot fine-grained classification performance. 
It contains three tasks: roof shape classification, road smoothness classification, and road surface material classification. 
We use the same data collection approach as the main SkyScript dataset except that the images are collected with Google and Bing Maps API instead of GEE. 
Also, objects that have already been included in SkyScript are excluded from this test set. 
These considerations ensure that both the image styles and objects in this test set are not seen during training. 
Class labels for every image in this test set are validated by human. 
Figure \ref{fig:fine_grained} shows examples of each class.

\subsection{A.8. Additional results}

\subsubsection{A.8.1. Trade-off between dataset size and data quality.}
\begin{figure*}[t]
\centering
\includegraphics[width=1\linewidth]{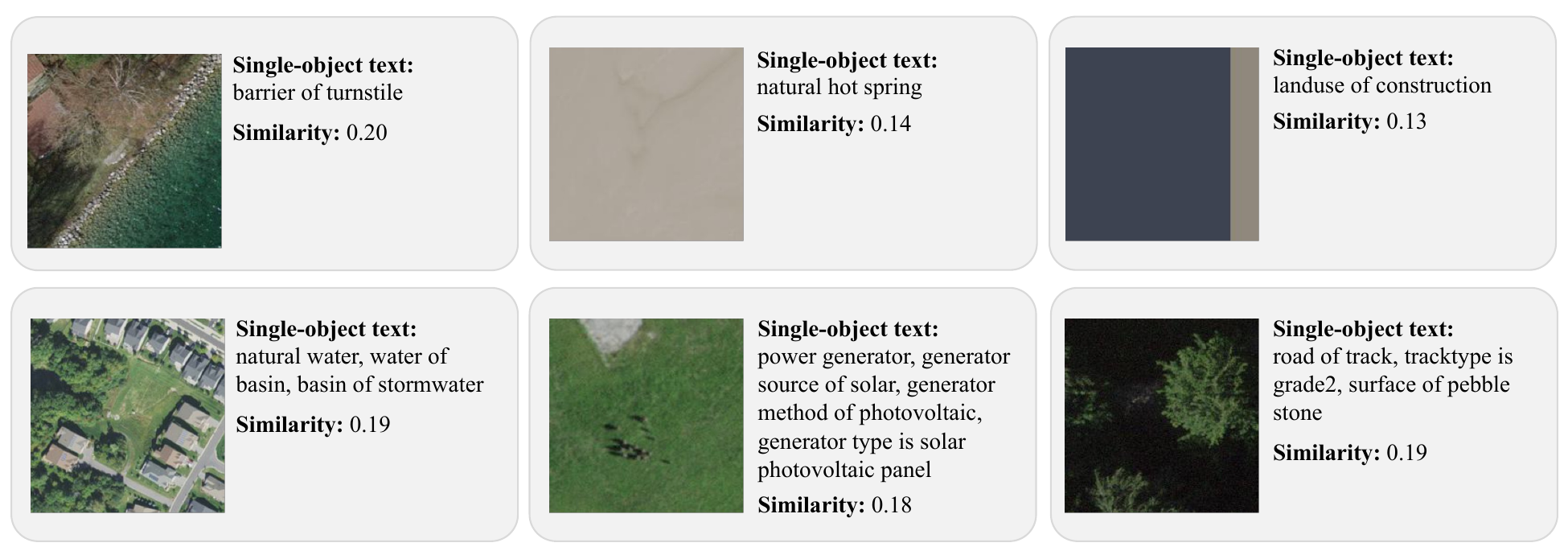}
\caption{Examples of image-text pairs with very low pairwise similarity. 
Typical causes of low similarity include: the image is not fully loaded; the object described by the caption is obscured by trees; the object described by the caption was constructed after the image capturing date; incorrect OSM annotation.
}
\label{fig:bad_examples}
\end{figure*}

Table \ref{tab:classification} shows how the average top-1 accuracy of scene classification of SkyCLIP—the CLIP model trained on SkyScript—varies with different subsets of SkyScript. 
Here SkyCLIP-X means only top X\% of image-text pairs in the full SkyScript dataset are selected for training, based on the descending order of the pairwise similarity. 
As it shows, by increasing the dataset size at the cost of including more low-similarity image-text pairs, the average accuracy increases. 
However, using the full dataset without any filtering yield a lower performance than using only 20\% high-quality data for training.
This indicates that the harm caused by including very-low-similarity image-text pairs outweighs the benefit of increasing dataset size. 
Examples of these very-low-similarity image-text pairs are shown in Figure \ref{fig:bad_examples}.

\begin{table}[h]
\centering
\resizebox{\columnwidth}{!}{
\begin{tabular}{lcc}
\toprule
Model  &   SkyScript dataset size &  Avg. top-1 accuracy (\%)  \\ 
\midrule
CLIP-original     & -         & 53.76      \\
SkyCLIP-20        & 1M        & 59.81      \\ 
SkyCLIP-30        & 1.5M      & 59.91      \\
SkyCLIP-50        & 2.6M      & \textbf{59.93}    \\
SkyCLIP-100       & 5.2M      & 58.65         \\

\bottomrule
\end{tabular}}
\caption{Average top-1 accuracy on seven scene classification benchmark datasets. ViT-L-14 is used as the CLIP image encoder. SkyCLIP-X means only top X\% of image-text pairs in the full SkyScript dataset are selected for training, based on the descending order of the pairwise similarity.}
\label{tab:classification}
\end{table}

\subsubsection{A.8.2. Full results on cross-modal retrieval.} 

Table \ref{tab:retrieval} shows recall@1, recall@5, and recall@10 for cross-modal retrieval on three benchmark datasets (RSICD, RSITMD, and UCM-caption). 
For AMFMN \cite{yuan2022exploring}, LW-MCR \cite{Yuan2022ALM}, and GaLR \cite{Yuan2022RemoteSC}, benchmark datasets are seen during training. 
For CLIP-based models, benchmark datasets are not seen during training. 
We find that, for most recall metrics, SkyCLIP surpasses the original CLIP as well as the CLIP model trained on LAION-RS, the remote sensing subset of LAION-2B. 
Furthermore, on RSICD and UCM-caption datasets, SkyCLIP can outperform the previous models that have been exposed to the benchmark datasets during training. 
On the RSITMD dataset, SkyCLIP's performance is comparable to the previous models but cannot consistently exceed them.
This demonstrate the value of SkyScript, which can facilitate the learning of versatile visual and text representations, enabling the zero-shot transfer to unseen cross-modal remote sensing retrieval tasks.

\subsubsection{A.8.3. Using remote-sensing-specialized CLIP models for filtering out uncorrelated pairs.} 
In the main paper, we use the original CLIP model (OpenAI's ViT-L/14 CLIP) for filtering out uncorrelated image-text pairs in the dataset.
Here we present additional results derived by using the CLIP-laion-RS model (continual pre-training with the remote sensing image subset of LAION-2B) to filter out uncorrelated image-text pairs, as is shown in Table \ref{tab:cls_results}. 
We find that using the remote-sensing-specialized CLIP (CLIP-laion-RS) for filtering can achieve slightly better performance than using the original CLIP for filtering (60.69\% vs. 59.91\% in terms of average accuracy in zero-shot scene classification).

\subsubsection{A.8.4. Performance on unseen classes.}
We perform a simple experiment to compare the performance of SkyCLIP with the original CLIP on the classes unseen during training. 
To do so, we exclude 10 classes (baseball field, basketball court, tennis court, substation, solar panel, wind turbine, oil field, wastewater plant, cemetery, airport) from SkyScript by filtering out relevant keywords.
We then use the remaining image-text pairs for training. 
The average accuracy of SkyCLIP on these unseen classes is 96.1\% on MillionAID and 94.0\% on PatternNet, (vs. 93.6\% and 86.5\% for CLIP). 
This shows the generalization cabability of the model developed with SkyScript on unseen classes.

\begin{table*}[htb]
    \resizebox{\linewidth}{!}{
        \begin{tabular}{lcccccccccccccccccc}
        \toprule
        \multirow{2}{*}{Model} & \multicolumn{6}{c}{RSICD} & \multicolumn{6}{c}{RSITMD} & \multicolumn{6}{c}{UCM-caption} \\
        \cmidrule(lr){2-7} \cmidrule(lr){8-13} \cmidrule(lr){14-19}
        & \multicolumn{3}{c}{Image to text} & \multicolumn{3}{c}{Text to image} & \multicolumn{3}{c}{Image to text} & \multicolumn{3}{c}{Text to image} & \multicolumn{3}{c}{Image to text} & \multicolumn{3}{c}{Text to image} \\
        \cmidrule(lr){2-4} \cmidrule(lr){5-7} \cmidrule(lr){8-10} \cmidrule(lr){11-13} \cmidrule(lr){14-16} \cmidrule(lr){17-19}
        & R@1 & R@5 & R@10 & R@1 & R@5 & R@10 & R@1 & R@5 & R@10 & R@1 & R@5 & R@10 & R@1 & R@5 & R@10 & R@1 & R@5 & R@10 \\
        \midrule
        \multicolumn{19}{l}{RSICD, RSITMD, and/or UCM-caption seen in training} \\
        \midrule
        AMFMN & 5.39 & 15.08 & 23.40 & 4.90 & 18.28 & 31.44 & 10.63 & 24.78 & 41.81 & \textbf{11.51} & 34.69 & \textbf{54.87} & 16.67 & 45.71 & 68.57 & 12.86 & 53.24 & 79.43 \\
        LW-MCR-u & 4.39 & 13.35 & 20.29 & 4.30 & 18.85 & 32.34 & 9.73 & 26.77 & 37.61 & 9.25 & 34.07 & 54.03 & 18.10 & 47.14 & 63.81 & 13.14 & 50.38 & 79.52 \\
        GaLR & 6.59 & 19.85 & 31.04 & 4.69 & 19.48 & 32.13 & \textbf{14.82} & 31.64 & 42.48 & 11.15 & \textbf{36.68} & 51.68 & - & - & - & - & - & - \\
        \midrule
        \multicolumn{19}{l}{RSICD, RSITMD, and/or UCM-caption not seen in training} \\
        \midrule
        CLIP-original & 6.59 & 20.68 & 31.75 & 3.62 & 14.28 & 23.63 & 10.18 & 30.31 & 42.04 & 8.31 & 24.96 & 39.03 & 37.62 & 78.10 & 89.52 & 28.12 & \textbf{64.99} & 77.19 \\
        CLIP-laion-RS & 8.42 & 23.70 & 35.86 & 5.81 & 19.49 & 30.25 & 13.72 & 32.08 & 44.91 & 10.57 & 31.48 & 46.96 & \textbf{39.52} & 79.52 & 90.00 & 29.71 & 62.60 & 80.37 \\
        SkyCLIP-30 & \textbf{8.97} & \textbf{24.15} & \textbf{37.97} & \textbf{5.85} & \textbf{20.53} & \textbf{33.53} & 11.73 & \textbf{33.19} & \textbf{47.35} & 10.19 & 32.47 & 49.08 & 38.57 & \textbf{84.29} & \textbf{93.81} & \textbf{31.83} & 64.19 & \textbf{81.96} \\
        \bottomrule
        \end{tabular}}
        \caption{Recall@1, recall@5, and recall@10 for cross-modal retrieval on three benchmark datasets.  
        SkyCLIP-30 is trained on SkyScript (top 30\% samples in terms of  pairwise similarity) using multi-object captions. CLIP-laion-RS is trained on the LAION-RS dataset.
        }
      \label{tab:retrieval}
\end{table*}

\begin{table*}[h]
\centering
\resizebox{\linewidth}{!}{
\begin{tabular}{llcccccccccccc}
\toprule
\multirow{2}{*}{ViT} & \multirow{2}{*}{Model} & \multicolumn{9}{c}{Scene classification} & \multicolumn{3}{c}{Fine-grained classification} \\
\cmidrule(lr){3-3} \cmidrule(lr){4-11} \cmidrule(lr){12-14}
& & SkyScript & AID & EuroSAT & fMoW & Million-AID & PatternNet & RESISC & RSI-CB & Avg. & Roof shape & Smoothness & Surface \\
\midrule
\multirow{2}{*}{L-14} 
& SkyCLIP-30 & 69.08 & 72.15 & 52.44 & \textbf{27.77} & 66.40 & 79.67 & 70.77 & 50.19 & 59.91 & 46.17 & 30.80 & 64.32 \\
& SkyCLIP-30-laion-filter & \textbf{69.97} & \textbf{72.50} & \textbf{55.15} & 27.61 & \textbf{66.86} & \textbf{80.24} & \textbf{70.97} & \textbf{51.50} & \textbf{60.69} & \textbf{47.33} & \textbf{38.00} & \textbf{66.36} \\ 
\bottomrule
\end{tabular}}
\caption{Top-1 accuracy (\%) for zero-shot scene classification and fine-grained classification. ``SkyCLIP-30-laion-filter'' denotes the model trained with the top 30\% most similar image-text pairs determined by CLIP-laion-RS (a CLIP model trained with the remote sensing image subset of LAION-2B).}
\label{tab:cls_results}
\end{table*}

\begin{figure*}[t]
\centering
\includegraphics[width=1\linewidth]{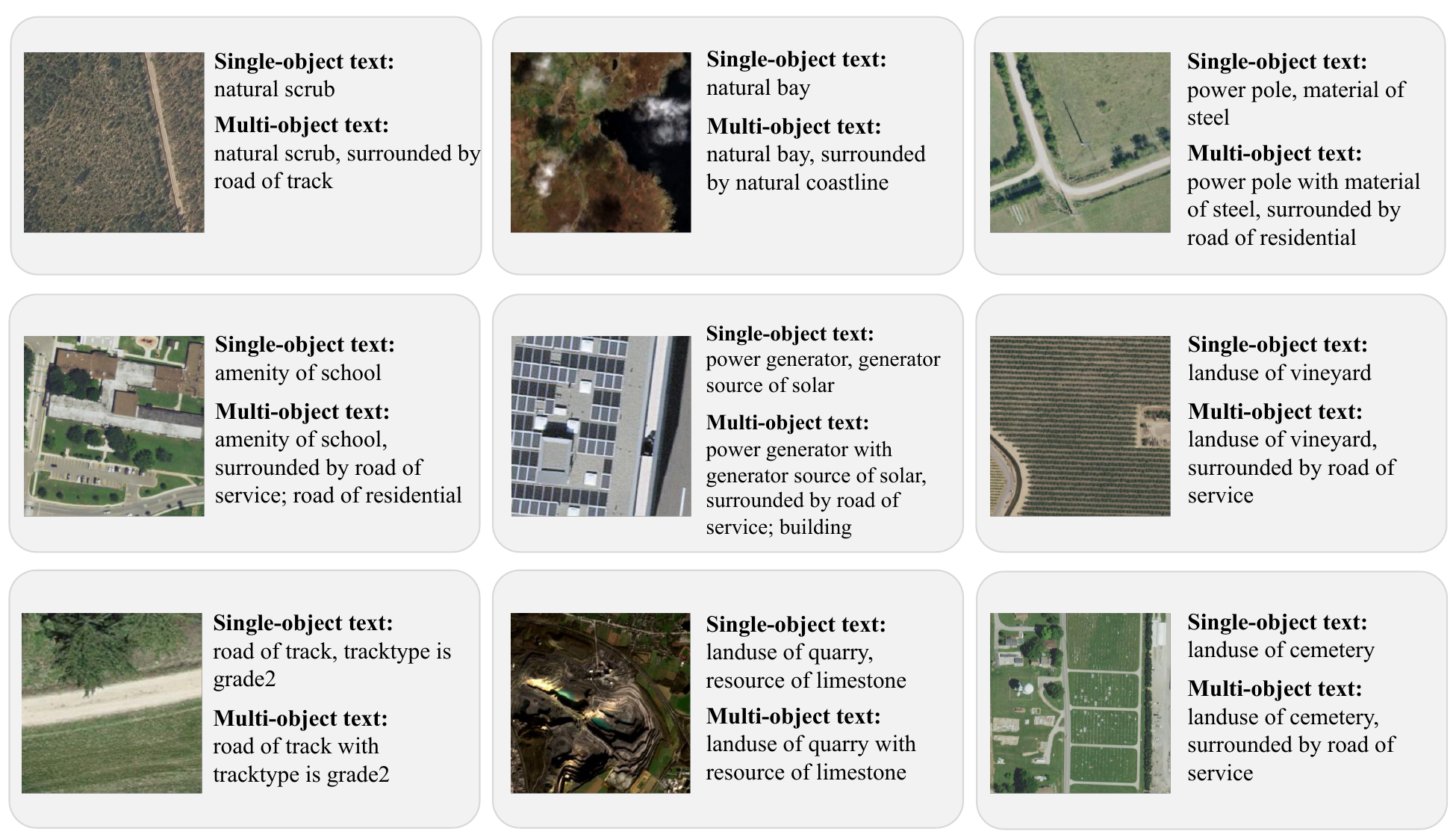}
\caption{Additional examples of image-caption pairs in the SkyScript dataset. Each image corresponds to a caption describing a single object and a caption describing multiple objects.
}
\label{fig:more_examples}
\end{figure*}